\pdfoutput=1

\documentclass[11pt]{article}

\usepackage[]{acl}
\usepackage{times}
\usepackage{latexsym}
\usepackage{graphicx}
\usepackage{float}
\usepackage{color, colortbl}
\definecolor{Gray}{gray}{0.9}
\usepackage{graphicx}
\usepackage{makecell}
\usepackage{lipsum}
\usepackage{float}
\usepackage{placeins}
\usepackage{hyperref}
\usepackage{epstopdf}
\usepackage{algorithm}
\usepackage{amsmath}
\usepackage{amssymb}
\usepackage{multirow}
\usepackage{subcaption}
\usepackage{orcidlink}

\usepackage[T1]{fontenc}

\usepackage[utf8]{inputenc}

\usepackage{microtype}

\title{ToxVidLM: A Multimodal Framework for Toxicity Detection in Code-Mixed Videos$^\dag$}

\author{Krishanu Maity$^{1, *}$, A.S. Poornash$^{1, *}$ \orcidlink{0009-0003-6335-9651} , Sriparna Saha$^{1}$ and Pushpak Bhattacharyya$^{2}$
\\$^{1}$Department of Computer Science and Engineering, Indian Institute of Technology Patna
\\$^{2}$Department of Computer Science and Engineering, Indian Institute of Technology Bombay\\
\texttt{\{krishanu\_2021cs19, poornash\_2101cs01, sriparna\}@iitp.ac.in, pb@cse.iitb.ac.in}
}

\begin{document}
\maketitle
\begin{abstract}

\let\thefootnote\relax\footnotetext{ $^*$ Denotes an equal contribution to this work by the respective authors and are jointly the first authors.}

\let\thefootnote\relax\footnotetext{ $^\dag$ This work has been accepted as a long paper at ACL 2024 Findings.}

In an era of rapidly evolving internet technology, the surge in multimodal content, including videos, has expanded the horizons of online communication. However, the detection of toxic content in this diverse landscape, particularly in low-resource code-mixed languages, remains a critical challenge. While substantial research has addressed toxic content detection in textual data, the realm of video content, especially in non-English languages, has been relatively underexplored. This paper addresses this research gap by introducing a benchmark dataset, the first of its kind, consisting of 931 videos with 4021 code-mixed Hindi-English utterances collected from YouTube. Each utterance within this dataset has been meticulously annotated for toxicity, severity, and sentiment labels. 
We have developed an advanced Multimodal Multitask framework built for {\bf Tox}icity detection in {\bf Vid}eo Content by leveraging Language Models (\textbf{LM}s), crafted for the primary objective along with the additional tasks of conducting sentiment and severity analysis. {\em ToxVidLM} incorporates three key modules – the Encoder module, Cross-Modal Synchronization module, and Multitask module – crafting a generic multimodal LM customized for intricate video classification tasks. Our experiments reveal that incorporating multiple modalities from the videos substantially enhances the performance of toxic content detection by achieving an Accuracy and Weighted F1 score of 94.29\% and 94.35\%, respectively. \footnote{The code and dataset will be made available at \href{https://github.com/justaguyalways/ToxVidLM_ACL_2024}{https://github.com/justaguyalways/ToxVidLM\_ACL\_2024}}\\

{\bf Disclaimer:} The article contains profanity, an inevitable situation for the nature of the work involved. These in no way reflect the opinion of the authors.
\end{abstract}

\section{Introduction}
In an age where social media platforms empower users to become content creators, the digital landscape has witnessed an unprecedented proliferation of information dissemination. By 2023, it is estimated that 82\% of internet traffic will be video content~\cite{WilsonA}. As a result, platforms like YouTube and Dailymotion have become major sources of information. A remarkable statistic underscores the colossal impact of these platforms: on YouTube alone, users collectively view more than a billion hours of video content each day\footnote{\href{https://blog.youtube/press/}{https://blog.youtube/press/}}. The viral nature of video content is a double-edged sword: it facilitates rapid news propagation yet simultaneously accelerates the dissemination of toxic speech. We adhere to the definition of toxic speech provided by~\citet{dixon2018measuring}, which characterizes it as {\em "discourteous, disrespectful, or unreasonable language likely to compel someone to exit a discussion".}

This expansive realm of videos on platforms like YouTube encompasses an array of topics, with the majority of content being innocuous. However, there exists a darker side – videos that blatantly contravene community guidelines and foster harmful narratives~\cite{blogBuildingGreater}. The non-removal of toxic content from these platforms can have severe repercussions, including the formation of hostile online environments with echo chambers of hateful users, potential loss of revenue, fines, and legal entanglements\footnote{\href{https://www.wsj.com/articles/germany-to-social-networks-delete-hate-speech-faster-or-face-fines-1498757679}{https://www.wsj.com/articles/germany-to-social-networks-delete-hate-speech-faster-or-face-fines-1498757679}}. While some platforms deploy human moderators to identify and remove harmful content, the sheer volume of daily user-generated content poses an overwhelming challenge. Facebook, for instance, engages approximately 15,000 moderators to review content flagged by both AI algorithms and users but still faces approximately 300,000 content moderation mistakes every day\footnote{\href{https://www.forbes.com/sites/johnkoetsier/2020/06/09/300000-facebook-content-moderation-mistakes-daily-report-says/?sh=777a39954d03}{https://www.forbes.com/sites/johnkoetsier}}. Furthermore, the toll on human moderators is not limited to their professional obligations but extends to the risk of emotional and psychological trauma. This issue is further compounded by legal regulations that mandate the swift removal of hateful content. Non-compliance with these laws could result in substantial fines.


Current research in the domain of toxic speech detection primarily focuses on text-based models~\cite{kennedy2020contextualizing,roy2022cyberbullying,obaid2023cyberbullying,maity2022emoji,das2022data,maity2023genex}, with limited exploration of image-based methodologies~\cite{yang2019exploring,gomez2020exploring,kiela2020hateful,maity2022multitask} and very few works on video data~\cite{wu2020detection,rana2022emotion,das2023hatemm,jha-etal-2024-meme}. 
Detecting harmful actions in videos requires the fusion of multi-frame video and speech processing signals, making direct adaptation of image-based hate detection methods inadequate. Existing toxic content detection methods predominantly rely on text-based modalities, with limited exploration of video content and a focus on monolingual languages like English. However, the surge in code-mixed language use, especially in multilingual countries like India, where people frequently blend Hindi and English in their communication (known as code-mixing~\cite{cm1}), presents a unique challenge for machine learning tool development, as highlighted by~\cite{vyas2014pos}. 

While studies have addressed toxic content detection in code-mixed social media texts, a significant research gap remains in code-mixed videos.

{\bf Main Contributions:}
This paper strives to address these challenges by introducing a comprehensive approach for detecting toxic speech in video content, leveraging the multi-modal nature of video data and advanced deep learning techniques. Through the development of efficient models, it aims to contribute to the creation of safer online environments and facilitate compliance with evolving legal regulations concerning toxic content. Our contributions are twofold:

i) We introduce {\em ToxCMM}, an openly accessible dataset extracted from YouTube that is meticulously annotated for toxic speech, with utterances presented in code-mixed form. Each sentence within the videos is annotated with three crucial labels, namely Toxic (Yes / No), Sentiment (Positive / Negative / Neutral), and Severity levels (Non-harmful / Partially Harmful / Very Harmful). This extensive dataset comprises 931 videos, encompassing a total of 4021 utterances. The release of the {\em ToxCMM} dataset is intended to foster further exploration in the realm of multi-modal toxic speech detection within low-resource code-mixed languages.

ii) We have innovated ToxVidLM, a multimodal multitask framework for detecting toxic videos and analyzing their sentiment and severity. ToxVidLM integrates three key modules: the Encoder module, the Cross-Modal Synchronization Module, and the Multitask module, to create a versatile Multimodal LM tailored for video classification tasks. Our framework incorporates a sophisticated gated modality fusion mechanism, empirically proven to outperform standard fusion techniques in ablation studies. We propose a method for synchronizing the text modality with other modalities, yielding promising results as demonstrated in our studies. Notably, our framework is adaptable to various publicly available pre-trained models, serving as modality encoders, making it applicable to diverse problem statements. Our most effective multitask model achieves notable weighted F1-Scores of 94.35\%, 86.84\%, and 83.42\% for Toxicity detection, Severity levels, and Sentiment identification, respectively.

\section{Related Works}
The widespread availability of multi-modal data has led to the utilization of multi-modal deep learning techniques, enhancing the accuracy of diverse tasks such as visual question answering~\cite{singh2019towards}, summarization~\cite{ghosh2024medsumm,ghosh2024clipsyntel} and the detection of fake news and rumors~\cite{khattar2019mvae}. In recent times, multi-modal hate speech detection has gained traction, where text posts are augmented with additional contextual information such as user and network data~\cite{founta2019unified} or images~\cite{yang2019exploring,gomez2020exploring,kiela2020hateful,maity2022multitask} to bolster detection accuracy. These multi-modal approaches often involve the utilization of unimodal methods like CNNs, LSTMs, or BERT for text encoding and deep CNNs like ResNet or InceptionV3 for image encoding. Subsequently, multi-modal fusion is performed through techniques like simple concatenation, gated summation, bilinear transformation, or attention-based methods. Additionally, the application of multi-modal transformers such as ViLBERT and Visual BERT has been explored~\cite{kiela2020hateful}.

While research on the detection of offensive or toxic videos is scarce, particularly in the context of languages such as Portuguese~\cite{alcantara2020offensive}, Thai~\cite{10494986} and English~\cite{wu2020detection,rana2022emotion,das2023hatemm}, it is worth noting that the existing work in this area predominantly revolves around monolingual languages. \citet{10494986} leverages textual data to propose a dual-channel network to classify hate and sentiment in low-resource settings. Notably, the two studies \cite{alcantara2020offensive,wu2020detection} exclusively considered textual features by extracting video transcripts for classification. In contrast,~\cite{rana2022emotion}'s research takes both textual and audio features for offensive video detection. However, this study confronts issues related to dataset accessibility, insufficiently detailed data curation and annotation processes, and a lack of precise dataset statistics. \citet{das2023hatemm} developed a more comprehensive approach by integrating all three modalities (text, Image, and audio) for hate video detection in English. 

To the best of our knowledge, our study pioneers the introduction of a multi-modal toxic video dataset in the context of low-resource code-mixed languages, further distinguished by the annotation of sentence-level labels. We are confident that our dataset, along with the benchmark models developed using it, will significantly assist content moderators in distinguishing genuine cases of hateful content while concurrently reducing false alarms.

\section{Toxic Code-Mixed Multimodal ({\em ToxCMM}) Dataset Creation}
\textbf{Data Collection:}
We selected YouTube as our primary data source, given its popularity as a video hosting platform. Our focus was on code-mixed language conversations, primarily in Hindi and English. To collect relevant content, we utilized the YouTube API to scrape Indian web series and Hindi "roasted" videos. We subdivided the downloaded videos into smaller sub-videos to annotate them at the sentence level and maximize the inclusion of toxic content. Initially, we obtained 1023 videos, but after a thorough review, we retained 931 videos as the remaining ones were mostly in English, not the intended Hindi-English code-mixed format. To generate transcripts for each video, we used the Whisper~\cite{radford2023robust} transcribing model, configured with word timestamps from the OpenAI library. We then manually improved the transcript quality by removing unclear words and symbols resulting from speech disruptions or stammering. Extracting individual utterances from the videos involved cataloging their start and end times.

\begin{table*}[hbt]
\caption{Class-wise data statistics of our developed {\em ToxCMM} dataset}
\label{tab:stat}
\scalebox{0.95}{
\begin{tabular}{c|c|cc|ccc|ccc}
\hline
\multirow{2}{*}{\textbf{\# Video}} & \multirow{2}{*}{\textbf{\#Utterances}} & \multicolumn{2}{c|}{\textbf{Toxicity}} & \multicolumn{3}{c|}{\textbf{Severity}} & \multicolumn{3}{c}{\textbf{Sentiment}}                   \\ \cline{3-10} 
                                   &                                        & \textbf{Non-Toxic}   & \textbf{Toxic}  & \textbf{0}  & \textbf{1}  & \textbf{2} & \textbf{Positive} & \textbf{Neutral} & \textbf{Negative} \\ \hline
931                                & 4021                                   & 2324                 & 1697            & 2324        & 834         & 863        & 469               & 1401              & 2151              \\ \hline
\end{tabular}}
\end{table*}

\begin{figure*}[t]
	\centering
        \includegraphics[width=\textwidth]{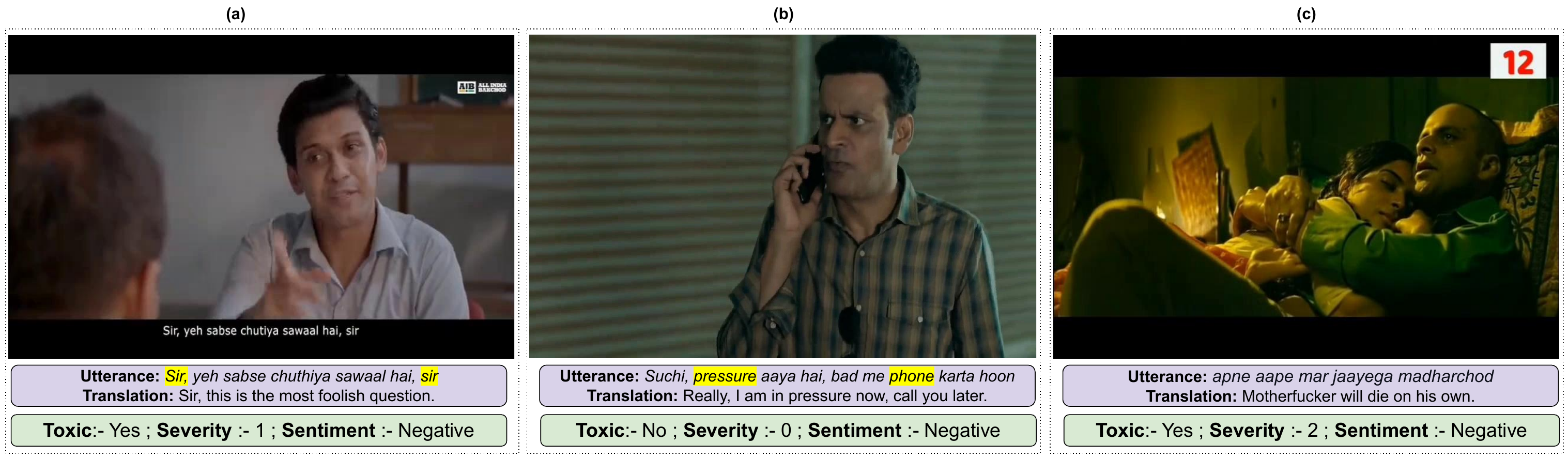}
	\caption {Some samples from annotated {\em ToxCMM} dataset; The yellow highlighted words are in English.}
	\label{tab:my-samples}
	
\end{figure*} 

\subsection{Data Annotation}
To better clarify the annotation process, we split the annotation section into two subsections: (i) Annotation Training and (ii) Main Annotation.\\
{\bf Annotation training: }
Three PhD scholars oversaw the annotation process, well-versed in toxic and offensive content, and the actual annotations were conducted by three undergraduate students proficient in both Hindi and English. Initially, we hired a group of masters students in linguistics who volunteered via our department email list and compensated them with gift vouchers and an honorarium. To train our annotators, we required gold standard samples with annotations for toxicity, severity, and sentiment labels. Our expert annotators randomly selected 150 samples (a small video of one sentence) and assigned suitable target classes. We considered two toxicity classes (Non-toxic/toxic), three sentiment classes (positive/neutral/negative), and the severity score on a three-point scale (0, 1, 2) for each video sample. Score 0 signifies that there is no indication of toxicity and 1 indicates that the post contains indications of toxicity. However, they are not severe, and a score of 2 indicates that the post contains strong evidence of toxicity (e.g., physical threats or excitement to commit suicide). Expert annotators engaged in discussions to resolve any differences and created 150 gold-standard samples with rationale and target annotations. These 150 annotated examples were divided into three sets, each containing 50 samples, to facilitate a three-phase training process. After each phase, expert annotators collaborated with novice annotators to correct any inaccuracies in the annotations, and the annotation guidelines were updated as needed. Following the conclusion of the third round of training, the top three annotators were selected to annotate the entire dataset containing 4021 samples.

{\bf Main annotation: }
We began with a small batch of 100 samples, gradually increasing it to 500 as annotators improved. To ensure consistency, we corrected errors from previous batches, and for final labels, majority voting was employed. In cases of disagreements among annotators, expert input was sought. Annotators were instructed to be unbiased in their assessments. The quality of annotations was assessed using Fleiss' Kappa scores~\cite{fleiss1971measuring}, resulting in IAA scores of 0.74 for toxicity classification, 0.67 for sentiment classification, and 0.64 for severity detection, confirming dataset quality and reliability. Figure~\ref{tab:my-samples} shows some samples from annotated {\em ToxCMM} dataset. Sample (b) shows a non-toxic video with negative sentiment. In contrast, both samples (a) and (c) are identified as toxic videos with negative sentiments. However, sample (a) is deemed less severe, while sample (c) is considered more severe due to its explicit use of profanity, aggressive language, and the wish for harm or death upon someone.

\subsection{Dataset Statistics }
The {\em ToxCMM} dataset comprises a total of 4021 utterances, with  1,697 categorized as toxic and the remaining 2,324 labelled as non-toxic. Class-wise statistics for the {\em ToxCMM} dataset are provided in Table \ref{tab:stat}. Each utterance in this dataset contains an average of 8.68 words, with an average duration of 8.89 seconds. On average, each utterance in the dataset contains about 68.20\% Hindi words, which means that more than two-thirds of the words are in Hindi, while the remaining words are in English.

\section{Methodology}
\textbf{Problem Formulation:}
We formulate our problem as follows: Given an utterance video clip denoted as V, our task is essentially a classification problem. We aim to determine whether the video contains toxic content, as well as assign sentiment and severity labels to it. Each video, $V$, is expressed as a sequence of frames,  $F =\left \{ f_{1}, f_{2}, \ldots f_{n}\right \}$, accompanied by its associated audio $A$ that is sampled at 16kHz to construct a sequence of features $A =\left \{ a_{1}, a_{2}, \ldots a_{l}\right \}$ and a transcript of the video,  $T =\left \{ w_{1}, w_{2}, \ldots w_{m}\right \}$, which consists of a sequence of words. Our goal is to construct a deep learning-based video classifier, denoted as $C : C(T;F;A) \rightarrow y$, where $y$ signifies the actual label of the video for a given task. 

\begin{figure*}[hbt]
	\centering
        \includegraphics[width=\textwidth]{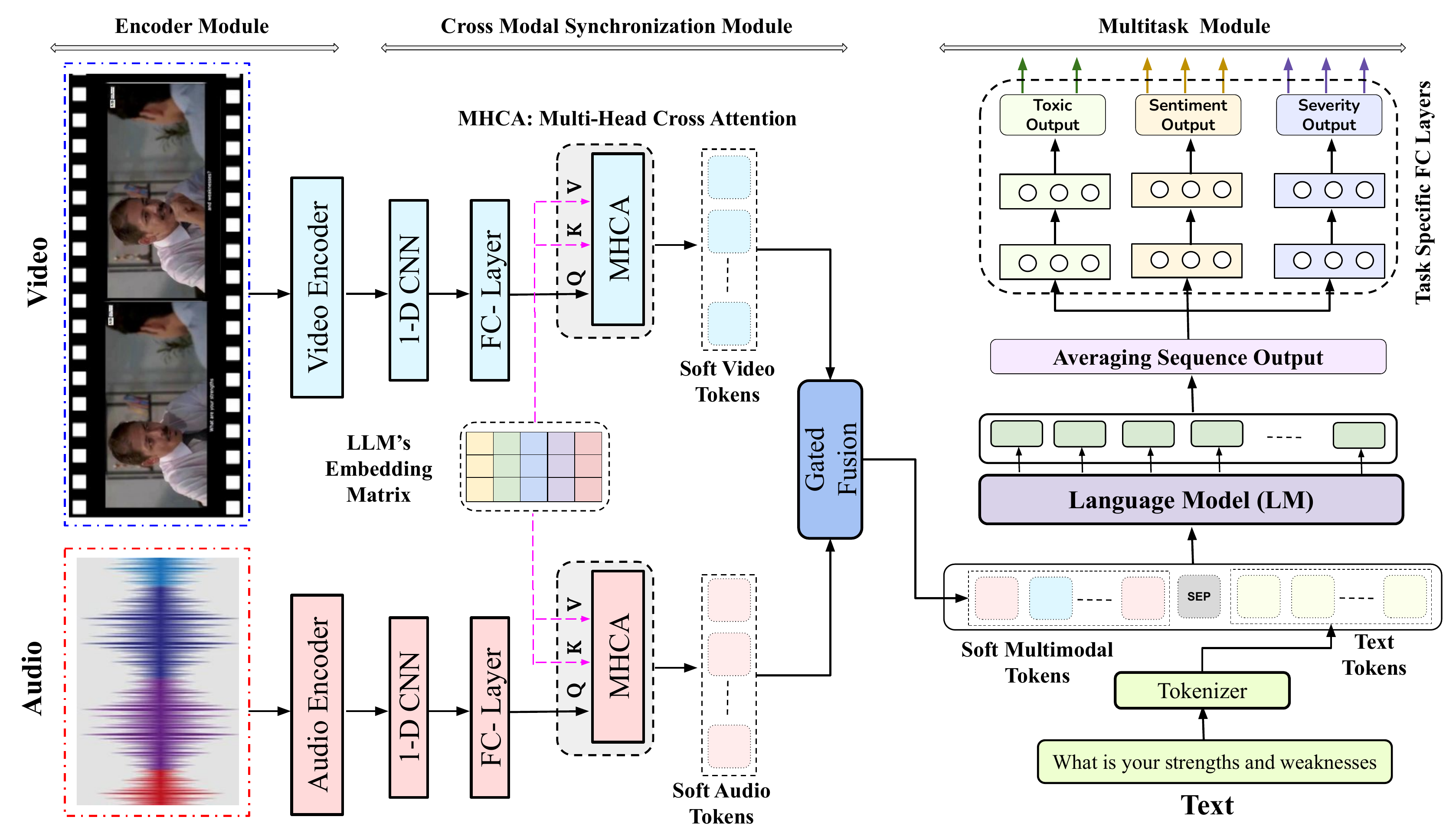}
	\caption{Architecture of proposed {\em ToxVidLM} model.}
	\label{fig:archi}
	
\end{figure*}

In this section, we describe our developed LM-based multimodal-multitask framework {\em ToxVidLM} (see Figure~\ref{fig:archi}) for toxic video detection and its sentiment and severity analysis. To enhance comprehension of our proposed method, we partition it into three distinct components: namely, the Encoder module, Cross Modal Synchronization Module, and the Multitask module.

\subsection{Encoder Module}
Current transformed-based language models (LMs) exhibit substantial capability but are commonly constrained to processing textual data exclusively. In this section, we elucidate our approach to encoding information from diverse modalities.

\textbf{Audio Encoder}
We conducted experiments using two state-of-the-art (SOTA) models, namely Whisper~\cite{radford2023robust} and MMS~\cite{pratap2023scaling}, to encode the audio signals and derive meaningful representations from the audio data. 

\textbf{Video Encoder:}
To handle the spatial and temporal information in
the videos, we consider two vision-based models, VideoMAE~\cite{tong2022videomae} and Timesformer~\cite{bertasius2021space}.

\textbf{Text Encoder:}
To generate text embeddings, we leverage the BERT~\cite{vaswani2017attention} family, renowned for its high effectiveness in various NLP tasks. Given our focus on Hindi-English code-mixed text, we have conducted experiments with three various models in HingBERT~\cite{nayak2022l3cube} family, like HingMBERT, HingRoBERTa, HingGPT and IndicBERT~\cite{kakwani2020indicnlpsuite}. These models are pre-trained on Hindi-English code-mixed Roman text. 

As per the results obtained, we utilize pre-trained models such as VideoMAE and Whisper to encode the input video ($V$) and audio ($A$) inputs, respectively, as outlined below:
\begin{subequations}
\begin{align}
    Z_{v} &= \text{VideoMAE}(V) \\
    Z_{a} &= \text{Whisper}(A)   
\end{align}
\end{subequations}
Also, $Z_{v} \in \mathbb{R}^{SL_{v} \times d_{v}}$ and $Z_{a} \in \mathbb{R}^{SL_{a} \times d_{a}}$ respectively. Here, $SL_{v}$ and $SL_{a}$ represent the sequence lengths of video and audio inputs, respectively.$d_{v}$ and $d_{a}$ represent the embedding dimension of encoded audio and video, respectively. Please see Appendix~\ref{sec:encoder} for more details on audio, video and text encoders used in our study.

\subsection{Cross Modal Synchronization Module}

Modality encoders are typically trained independently, resulting in discrepancies among the generated representations. Consequently, it becomes imperative to synchronize these distinct representations within a unified space to enhance the overall coherence and effectiveness of multimodal processing. In this section, we present a detailed methodology for aligning these representations.

{\bf Modality Synchronization}
The synchronization strategy aims to effectively correlate features extracted from multiple modalities, such as audio and video, with a primary focus on textual features. This emphasis is due to the heightened significance of textual information in addressing our specific problem statement. Textual features are preferred over auditory and visual signals due to their comparatively lower susceptibility to noise. This empirical observation is consistently reflected in the obtained results, emphasizing the importance of robust textual representation. 

The procedure for modality synchronization is delineated as follows:

{\textbf{(1) Abstract Feature Extraction:}} To mitigate computational expenses and limit the token count in the prefix, we utilize a 1-D convolutional layer (\textit{Conv}) to compress the length of multi-modal features to a condensed and consistent value. Following this, a linear layer (\textit{FC}) is employed to modify the hidden size of the features, aligning it with the dimensions of the token embeddings in the LMs, as described below:
\begin{subequations}
\begin{align}
    C_{v}=FC(Conv(Z_{v})) \\
    C_{a}=FC(Conv(Z_{a})) 
\end{align}
\end{subequations}
where  $C_{v} \in \mathbb{R}^{SL^{'} \times d_{t}}$ and $C_{a} \in \mathbb{R}^{SL^{'} \times d_{t}}$ are the abstract features with a fixed length of $SL^{'}$. Here, $d_{t}$ represents the dimensionality of the embedding matrix $E_{llm} \in \mathbb{R}^{V \times d_{t}}$ associated with the textual LMs (i.e., HingRoberta or HingGPT in our work) with vocabulary size V.\newline{\textbf{(2) Multi-Head Cross Attention (MHCA)}}
To establish a unified representation space guided by the text modality, we implemented Multi-Head Cross-Attention (MHCA) on the abstract video and audio features obtained from the preceding layer. MHCA involves scaled dot-product attention applied to three inputs: Query (Q) from one modality, Key (K) and Value (V) from another. This attention mechanism computes attention weights by comparing the queries Q with the keys K, updating the query representations via a weighted sum of the values V, as described below:
\begin{equation}
\label{eq:qkv}
MHCA(Q,K,V) = Softmax\left ( \frac{QK\;^{T}}{\sqrt{d_{k}}} \right )V
\end{equation}
where $d_{k}$ is the dimensionality of the key and query vectors. 
Using the attention mechanism in Equation~\ref{eq:qkv}, we propose to align the audio and visual representations with the textual embedding space as follows:
\begin{equation}
\label{soft_tokens}
\begin{split}
C_{v}^{s}=MHCA(C_{v},E_{t},E_{t}),\\
C_{a}^{s}=MHCA(C_{a},E_{t},E_{t})
\end{split}
\end{equation}
In our research, we view the attenuated representations of visual ($C_{v}^{s}$) and audio ($C_{a}^{s}$) modalities derived from Equation \ref{soft_tokens} as the soft tokens utilized by the LM, acting as the input to the Multitask module.

{\textbf{(3) Gated Fusion}}
To combine soft video and audio tokens, we employ a gated fusion strategy. Unlike concatenation or directly assigning weights to each vector, the gate fusion mechanism enables varying contributions to the prediction from different positions of vectors. The joint representation resulting from the gate fusion is computed as follows:
\begin{equation}
\begin{split}
\alpha = \sigma  (\mathbb{P}_{v}C_{v}^{s} \;+ \;\mathbb{P}_{a}C_{a}^{s} \;+\; b_{g}),\\
J_{va} = \alpha \;C_{a}^{s} \;+\; (1-\alpha )C_{v}^{s}
\end{split}
\end{equation}
Here, $\mathbb{P}_{v}$ and $\mathbb{P}_{a}$ represent weight matrices for the visual and acoustic modalities, while $b_{g}$ denotes scalar bias and $\sigma$ is the sigmoid activation function.

\subsection{Multitask Module} Typically, the LM model processes the input transcript $T$, generating a text token embedding $E_{t}$ with dimensions $SL_{t} \times d_{t}$. Here, $SL_{t}$ is the maximum sequence length of the transcript, and $d_{t}$ is the embedding dimension. Here we have added the joint multimodal soft tokens ($J_{va}$) obtained from the Cross-Modal Synchronization Module, appended with the text tokens separated by the special token [SEP], thereby creating a multimodal input to the LM for a more comprehensive understanding of the input video. Subsequently, the sequence output from the LM undergoes averaging and is passed through three task-specific fully connected layers, followed by an output softmax layer, facilitating the concurrent solution of three tasks: toxicity, severity, and sentiment detection from a video.  
\subsection{Loss Function} 
The loss function used in all tasks is categorical cross-entropy. The final loss function ($Loss_{f}$) is a weighted sum of individual task-specific losses ($Loss_{s}$) for $M$ tasks, where the contribution of each task's loss to the overall loss is determined by the loss weight $\beta$ as shown in Equation~\eqref{eq:lossfn}. 
\begin{equation}
Loss_{f} = \sum_{k=1}^{M}\beta_{k}Loss_{s}^{k}
\label{eq:lossfn}
\end{equation}
Where the parameters $\beta_{i}$ are learnt end-to-end, signifying task contribution from task $i$ to the multitask loss, enabling differential importance for parameter updates across tasks.

\section{Experimental Results and Analysis}
\textbf{Experimental Settings: }
All experiments were conducted on a machine equipped with an Intel Xeon Gold 5218 CPU featuring 64 cores and 128 threads, coupled with four Nvidia Tesla V100 GPUs with VRAM memory of 30 GB per GPU card. For the experiments' preparation, the dataset was partitioned into testing, validation, and training sets at ratios of 10\%, 10\%, and 80\%, respectively. To ensure robustness, the models were trained ten times with different random splits, and the average performance was reported. Several network configurations were tested, with the best results achieved using the Adam optimizer \cite{kingma2014adam} with a Cosine Annealing Learning Rate scheduler \cite{loshchilov2016sgdr}, batch size of 2, a learning rate set to $1e^{-5}$, and training for 30 epochs.
All models were implemented in the PyTorch framework\footnote{\href{https://pytorch.org/}{https://pytorch.org/}.}.

\begin{figure}[hbt]
	\centering
	\includegraphics[height = 3.5 cm, width = 7 cm]{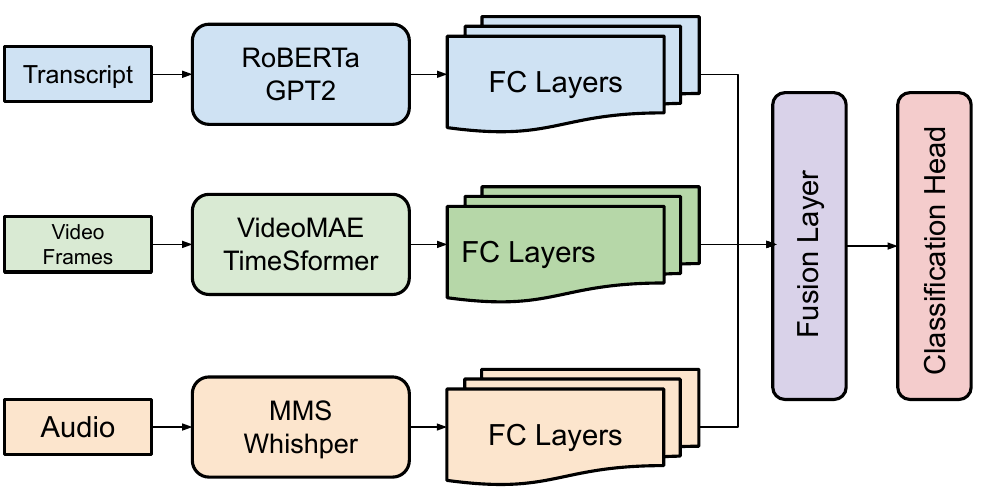}
	\caption {A schematic of baselines models as mentioned in~\cite{das2023hatemm}}
	\label{fig:baseline}
	
\end{figure}

{\bf Baseline Setup}
We have implemented the baseline model as outlined in the study by~\citet{das2023hatemm} which introduced a multimodal dataset designed for Hate-speech classification. In their proposed architecture, each modality is processed separately through a transformer-based encoder, followed by modality-specific fully connected layers. Subsequently, a Fusion Layer concatenates the modality-specific representations, which are then forwarded to a fully connected classification layer. The output dimension of this final layer corresponds to the number of classes in the classification task. The entire model is trained using a single cross-entropy loss function. We have exclusively conducted single-task experiments for the three tasks (Toxicity detection, Severity detection, and Sentiment classification) in unimodal/bimodal/trimodal settings. A diagram illustrating baseline models is presented in Figure~\ref{fig:baseline}

\begin{table}[hbt]
    \centering
\caption{
The outcomes of various transformer-based baseline models are presented within different modalities, i.e., Video (V), Audio (A), and Text (T) configurations for three tasks (Toxicity, Severity, and Sentiment). The results are measured in terms of weighted-average-F1 score (F1) and Accuracy (Acc) values. Bold-faced values represent the maximum scores attained;  TF - Timesformer, VM - VideoMAE, WP - Whisper, M-BERT - HingMBERT, GPT2 - HingGPT, RT - HingRoberta.}
\label{tab:ml_results}
\scalebox{0.65}{
\begin{tabular}{clcccccc}
\hline
\multicolumn{1}{l|}{}                                                                                     & \multicolumn{1}{l|}{}                                          & \multicolumn{2}{c|}{\textbf{Toxic}}                                                                  & \multicolumn{2}{c|}{\textbf{Severity}}                                                               & \multicolumn{2}{c}{\textbf{Sentiment}}                                          \\ \cline{3-8} 
\multicolumn{1}{l|}{\multirow{-2}{*}{\textbf{M}}}                                                         & \multicolumn{1}{l|}{\multirow{-2}{*}{\textbf{Model}}}          & \textbf{F1}                           & \multicolumn{1}{c|}{\textbf{Acc}}                            & \textbf{F1}                           & \multicolumn{1}{c|}{\textbf{Acc}}                            & \textbf{F1}                           & \textbf{Acc}                            \\ \hline
\multicolumn{8}{c}{\textbf{Baselines : Unimodal}}                                                                                                                                                                                                                                                                                                                                                                                                                          \\ \hline
\multicolumn{1}{c|}{}                                                                                     & \multicolumn{1}{l|}{TF}                                        & 66.23                                  & \multicolumn{1}{c|}{66.47}                                  & 60.38                                  & \multicolumn{1}{c|}{62.88}                                  & 57.12                                  & 58.24                                  \\
\multicolumn{1}{c|}{\multirow{-2}{*}{\textbf{V}}}                                                         & \multicolumn{1}{l|}{VM}                                        & 68.67                                  & \multicolumn{1}{c|}{68.73}                                  & 61.42                                  & \multicolumn{1}{c|}{64.26}                                  & 58.69                                  & 60.79                                  \\ \hline
\multicolumn{1}{c|}{}                                                                                     & \multicolumn{1}{l|}{MMS}                                       & 75.81                                  & \multicolumn{1}{c|}{75.89}                                  & 67.21                                  & \multicolumn{1}{c|}{67.35}                                  & 64.28                                  & 64.14                                  \\
\multicolumn{1}{c|}{\multirow{-2}{*}{\textbf{A}}}                                                         & \multicolumn{1}{l|}{WP}                                        & 76.18                                  & \multicolumn{1}{c|}{76.17}                                  & 68.97                                  & \multicolumn{1}{c|}{67.99}                                  & 65.45                                  & 65.01                                  \\ \hline
\multicolumn{1}{c|}{}                                                                                     & \multicolumn{1}{l|}{I-BERT}                                    & \multicolumn{1}{l}{81.72}              & \multicolumn{1}{l|}{81.86}                                  & \multicolumn{1}{l}{73.22}              & \multicolumn{1}{l|}{73.44}                                  & \multicolumn{1}{l}{68.25}              & \multicolumn{1}{l}{68.48}              \\
\multicolumn{1}{c|}{}                                                                                     & \multicolumn{1}{l|}{M-BERT}                                    & \multicolumn{1}{l}{83.28}              & \multicolumn{1}{l|}{83.23}                                  & \multicolumn{1}{l}{75.35}              & \multicolumn{1}{l|}{76.17}                                  & \multicolumn{1}{l}{71.26}              & \multicolumn{1}{l}{71.15}              \\
\multicolumn{1}{c|}{}                                                                                     & \multicolumn{1}{l|}{GPT2}                                      & 85.67                                  & \multicolumn{1}{c|}{85.71}                                  & 75.02                                  & \multicolumn{1}{c|}{75.93}                                  & 73.11                                  & 72.72                                  \\
\multicolumn{1}{c|}{\multirow{-4}{*}{\textbf{T}}}                                                         & \multicolumn{1}{l|}{\cellcolor[HTML]{EFEFEF}\textbf{RT}}       & \cellcolor[HTML]{EFEFEF}\textbf{86.98} & \multicolumn{1}{c|}{\cellcolor[HTML]{EFEFEF}\textbf{86.95}} & \cellcolor[HTML]{EFEFEF}\textbf{77.23} & \multicolumn{1}{c|}{\cellcolor[HTML]{EFEFEF}\textbf{76.42}} & \cellcolor[HTML]{EFEFEF}\textbf{73.41} & \cellcolor[HTML]{EFEFEF}\textbf{74.44} \\ \hline
\multicolumn{8}{c}{\textbf{Baselines : Bimodal}}                                                                                                                                                                                                                                                                                                                                                                                                                           \\ \hline
\multicolumn{1}{c|}{}                                                                                     & \multicolumn{1}{l|}{TF + MMS}                                  & 76.49                                  & \multicolumn{1}{c|}{76.53}                                  & 68.74                                  & \multicolumn{1}{c|}{68.89}                                  & 64.31                                  & 64.24                                  \\
\multicolumn{1}{c|}{}                                                                                     & \multicolumn{1}{l|}{VM + MMS}                                  & 77.25                                  & \multicolumn{1}{c|}{77.48}                                  & 69.42                                  & \multicolumn{1}{c|}{69.28}                                  & 65.18                                  & 65.49                                  \\
\multicolumn{1}{c|}{}                                                                                     & \multicolumn{1}{l|}{TF + WP}                                   & 78.72                                  & \multicolumn{1}{c|}{78.94}                                  & 70.34                                  & \multicolumn{1}{c|}{70.57}                                  & 67.85                                  & 67.41                                  \\
\multicolumn{1}{c|}{\multirow{-4}{*}{\textbf{\begin{tabular}[c]{@{}c@{}}V \\ +\\  A\end{tabular}}}}       & \multicolumn{1}{l|}{VM + WP}                                   & 79.23                                  & \multicolumn{1}{c|}{79.41}                                  & 71.79                                  & \multicolumn{1}{c|}{72.71}                                  & 69.45                                  & 69.47                                  \\ \hline
\multicolumn{1}{c|}{}                                                                                     & \multicolumn{1}{l|}{GPT2 + TF}                                 & 85.87                                  & \multicolumn{1}{c|}{85.84}                                  & 75.13                                  & \multicolumn{1}{c|}{75.91}                                  & 73.12                                  & 73.91                                  \\
\multicolumn{1}{c|}{}                                                                                     & \multicolumn{1}{l|}{GPT2 + VM}                                 & 86.71                                  & \multicolumn{1}{c|}{86.84}                                  & 76.12                                  & \multicolumn{1}{c|}{76.98}                                  & 74.03                                  & 74.45                                  \\
\multicolumn{1}{c|}{}                                                                                     & \multicolumn{1}{l|}{RT + TF}                                   & 87.01                                  & \multicolumn{1}{c|}{87.93}                                  & 76.92                                  & \multicolumn{1}{c|}{77.19}                                  & 73.92                                  & 74.56                                  \\
\multicolumn{1}{c|}{\multirow{-4}{*}{\textbf{\begin{tabular}[c]{@{}c@{}}T \\ + \\ V\end{tabular}}}}       & \multicolumn{1}{l|}{RT + VM}                                   & 87.11                                  & \multicolumn{1}{c|}{87.08}                                  & 77.68                                  & \multicolumn{1}{c|}{78.11}                                  & 74.11                                  & 74.59                                  \\ \hline
\multicolumn{1}{c|}{}                                                                                     & \multicolumn{1}{l|}{GPT2 + MMS}                                & 86.27                                  & \multicolumn{1}{c|}{86.35}                                  & 75.03                                  & \multicolumn{1}{c|}{76.21}                                  & 73.15                                  & 72.87                                  \\
\multicolumn{1}{c|}{}                                                                                     & \multicolumn{1}{l|}{GPT2 + WP}                                 & 86.77                                  & \multicolumn{1}{c|}{86.84}                                  & 75.73                                  & \multicolumn{1}{c|}{76.42}                                  & 74.06                                  & 73.94                                  \\
\multicolumn{1}{c|}{}                                                                                     & \multicolumn{1}{l|}{RT + MMS}                                  & 87.18                                  & \multicolumn{1}{c|}{87.19}                                  & 77.22                                  & \multicolumn{1}{c|}{78.15}                                  & 74.08                                  & 74.69                                  \\
\multicolumn{1}{c|}{\multirow{-4}{*}{\textbf{\begin{tabular}[c]{@{}c@{}}T \\ + \\ A\end{tabular}}}}       & \multicolumn{1}{l|}{\cellcolor[HTML]{EFEFEF}\textbf{RT + WP}}  & \cellcolor[HTML]{EFEFEF}\textbf{87.26} & \multicolumn{1}{c|}{\cellcolor[HTML]{EFEFEF}\textbf{87.33}} & \cellcolor[HTML]{EFEFEF}\textbf{77.42} & \multicolumn{1}{c|}{\cellcolor[HTML]{EFEFEF}\textbf{78.41}} & \cellcolor[HTML]{EFEFEF}\textbf{74.22} & \cellcolor[HTML]{EFEFEF}\textbf{74.93} \\ \hline
\multicolumn{8}{c}{\textbf{Baselines : Trimodal}}                                                                                                                                                                                                                                                                                                                                                                                                                          \\ \hline
\multicolumn{1}{c|}{}                                                                                     & \multicolumn{1}{l|}{GPT2+TF+MMS}                               & 86.92                                  & \multicolumn{1}{c|}{87.02}                                  & 75.85                                  & \multicolumn{1}{c|}{76.61}                                  & 74.27                                  & 74.69                                  \\
\multicolumn{1}{c|}{}                                                                                     & \multicolumn{1}{l|}{GPT2+VM+MMS}                               & 86.72                                  & \multicolumn{1}{c|}{87.12}                                  & 75.72                                  & \multicolumn{1}{c|}{76.48}                                  & 74.23                                  & 74.87                                  \\
\multicolumn{1}{c|}{}                                                                                     & \multicolumn{1}{l|}{GPT2+TF+WP}                                & 86.88                                  & \multicolumn{1}{c|}{87.27}                                  & 75.89                                  & \multicolumn{1}{c|}{76.64}                                  & 74.31                                  & 74.75                                  \\
\multicolumn{1}{c|}{}                                                                                     & \multicolumn{1}{l|}{GPT2+VM+WP}                                & 87.21                                  & \multicolumn{1}{c|}{87.34}                                  & 77.14                                  & \multicolumn{1}{c|}{77.17}                                  & 74.43                                  & 74.78                                  \\
\multicolumn{1}{c|}{}                                                                                     & \multicolumn{1}{l|}{RT+TF+MMS}                                 & 87.29                                  & \multicolumn{1}{c|}{87.39}                                  & 77.93                                  & \multicolumn{1}{c|}{78.47}                                  & 75.09                                  & 75.65                                  \\
\multicolumn{1}{c|}{}                                                                                     & \multicolumn{1}{l|}{RT+VM+MMS}                                 & 87.58                                  & \multicolumn{1}{c|}{87.79}                                  & 77.96                                  & \multicolumn{1}{c|}{78.72}                                  & 75.03                                  & 75.49                                  \\
\multicolumn{1}{c|}{}                                                                                     & \multicolumn{1}{l|}{RT+TF+WP}                                  & 87.82                                  & \multicolumn{1}{c|}{87.67}                                  & 78.16                                  & \multicolumn{1}{c|}{78.36}                                  & 75.11                                  & 75.66                                  \\
\multicolumn{1}{c|}{\multirow{-8}{*}{\textbf{\begin{tabular}[c]{@{}c@{}}T\\ +\\ V\\ +\\ A\end{tabular}}}} & \multicolumn{1}{l|}{\cellcolor[HTML]{EFEFEF}\textbf{RT+VM+WP}} & \cellcolor[HTML]{EFEFEF}\textbf{88.09} & \multicolumn{1}{c|}{\cellcolor[HTML]{EFEFEF}\textbf{88.08}} & \cellcolor[HTML]{EFEFEF}\textbf{78.19} & \multicolumn{1}{c|}{\cellcolor[HTML]{EFEFEF}\textbf{78.66}} & \cellcolor[HTML]{EFEFEF}\textbf{75.78} & \cellcolor[HTML]{EFEFEF}\textbf{75.82} \\ \hline
\end{tabular}}
\end{table}

\begin{table}[hbt]
\centering
\caption{
Results of proposed {\em ToxVidLM} framework with different modality configurations for three tasks (Toxicity,  Severity and Sentiment classification) in single and multitask settings; TF - Timesformer, VM - VideoMAE, WP - Whisper, M-BERT - HingMBERT, GPT2 - HingGPT, RT - HingRoberta. }
\label{tab:dl_results}
\scalebox{0.60}{
\begin{tabular}{clllcccc}
\hline
\multicolumn{1}{l|}{}                                    & \multicolumn{1}{l|}{}                                              & \multicolumn{2}{c|}{\textbf{Toxicity}}                                                                             & \multicolumn{2}{c|}{\textbf{Severity}}                                                                                                 & \multicolumn{2}{c}{\textbf{Sentiment}}                                                                                  \\ \cline{3-8} 
\multicolumn{1}{l|}{\multirow{-2}{*}{\textbf{Modality}}} & \multicolumn{1}{l|}{\multirow{-2}{*}{\textbf{Encoder}}}            & \multicolumn{1}{c}{\textbf{F1}}                  & \multicolumn{1}{c|}{\textbf{Acc}}                            & \textbf{F1}                                               & \multicolumn{1}{c|}{\textbf{Acc}}                                          & \textbf{F1}                                               & \textbf{Acc}                                                \\ \hline
\multicolumn{8}{c}{\textbf{Single Task}}                                                                                                                                                                                                                                                                                                                                                                                                                                                                           \\ \hline
\multicolumn{1}{c|}{}                                    & \multicolumn{1}{l|}{GPT2 + VM}                                     & \multicolumn{1}{c}{{\color[HTML]{374151} 90.15}}  & \multicolumn{1}{c|}{{\color[HTML]{374151} 90.09}}           & {\color[HTML]{374151} 83.35}                               & \multicolumn{1}{c|}{{\color[HTML]{374151} 82.99}}                         & {\color[HTML]{374151} 75.64}                               & {\color[HTML]{374151} 75.26}                               \\
\multicolumn{1}{c|}{\multirow{-2}{*}{\textbf{T+V}}}      & \multicolumn{1}{l|}{RT + VM}                                       & \multicolumn{1}{c}{{\color[HTML]{374151} 91.11}}  & \multicolumn{1}{c|}{{\color[HTML]{374151} 91.03}}           & {\color[HTML]{374151} 83.38}                               & \multicolumn{1}{c|}{83.87}                                                & 80.46                                                      & 80.09                                                      \\ \hline
\multicolumn{1}{c|}{}                                    & \multicolumn{1}{l|}{GPT2 + WP}                                     & \multicolumn{1}{c}{91.27}                         & \multicolumn{1}{c|}{91.06}                                  & 84.49                                                      & \multicolumn{1}{c|}{83.89}                                                & 76.61                                                      & 76.19                                                      \\
\multicolumn{1}{c|}{\multirow{-2}{*}{\textbf{T+A}}}      & \multicolumn{1}{l|}{RT + WP}                                       & \multicolumn{1}{c}{92.64}                         & \multicolumn{1}{c|}{92.58}                                  & 84.46                                                      & \multicolumn{1}{c|}{85.06}                                                & 81.16                                                      & 81.27                                                      \\ \hline
\multicolumn{1}{c|}{}                                    & \multicolumn{1}{l|}{GPT2 + VM+ WP}                                 & \multicolumn{1}{c}{92.14}                         & \multicolumn{1}{c|}{91.98}                                  & 84.86                                                      & \multicolumn{1}{c|}{84.92}                                                & 78.14                                                      & 78.23                                                      \\
\multicolumn{1}{c|}{\multirow{-2}{*}{\textbf{T+V+A}}}    & \multicolumn{1}{l|}{\cellcolor[HTML]{EFEFEF}RT + VM + WP}          & \multicolumn{1}{c}{\cellcolor[HTML]{EFEFEF}93.85} & \multicolumn{1}{c|}{\cellcolor[HTML]{EFEFEF}93.64}          & \cellcolor[HTML]{EFEFEF}{\color[HTML]{374151} 86.28}       & \multicolumn{1}{c|}{\cellcolor[HTML]{EFEFEF}{\color[HTML]{374151} 86.51}} & \cellcolor[HTML]{EFEFEF}{\color[HTML]{374151} 82.76}       & \cellcolor[HTML]{EFEFEF}{\color[HTML]{374151} 82.87}       \\ \hline
\multicolumn{8}{c}{\textbf{Multi-Task (Toxic + Severity)}}                                                                                                                                                                                                                                                                                                                                                                                                                                                         \\ \hline
\multicolumn{1}{c|}{}                                    & \multicolumn{1}{l|}{GPT2 + VM}                                     & 91.53                                             & \multicolumn{1}{l|}{91.31}                                  & \multicolumn{1}{l}{84.68}                                  & \multicolumn{1}{l|}{84.19}                                                & -                                                          & -                                                          \\
\multicolumn{1}{c|}{\multirow{-2}{*}{\textbf{T+V}}}      & \multicolumn{1}{l|}{RT + VM}                                       & 92.07                                             & \multicolumn{1}{l|}{92.38}                                  & \multicolumn{1}{l}{84.88}                                  & \multicolumn{1}{l|}{85.56}                                                & -                                                          & -                                                          \\ \hline
\multicolumn{1}{c|}{}                                    & \multicolumn{1}{l|}{GPT2 + WP}                                     & 92.47                                             & \multicolumn{1}{l|}{92.28}                                  & \multicolumn{1}{l}{85.79}                                  & \multicolumn{1}{l|}{85.28}                                                & -                                                          & -                                                          \\
\multicolumn{1}{c|}{\multirow{-2}{*}{\textbf{T+A}}}      & \multicolumn{1}{l|}{RT + WP}                                       & 93.61                                             & \multicolumn{1}{l|}{93.58}                                  & \multicolumn{1}{l}{85.87}                                  & \multicolumn{1}{l|}{86.43}                                                & -                                                          & -                                                          \\ \hline
\multicolumn{1}{c|}{}                                    & \multicolumn{1}{l|}{GPT2 + VM+ WP}                                 & 93.48                                             & \multicolumn{1}{l|}{93.29}                                  & \multicolumn{1}{l}{86.54}                                  & \multicolumn{1}{l|}{86.12}                                                & -                                                          & -                                                          \\
\multicolumn{1}{c|}{\multirow{-2}{*}{\textbf{T+V+A}}}    & \multicolumn{1}{l|}{\cellcolor[HTML]{EFEFEF}RT + VM + WP}          & \cellcolor[HTML]{EFEFEF}94.12                     & \multicolumn{1}{l|}{\cellcolor[HTML]{EFEFEF}93.87}          & \multicolumn{1}{l}{\cellcolor[HTML]{EFEFEF}86.56}          & \multicolumn{1}{l|}{\cellcolor[HTML]{EFEFEF}86.82}                        & \cellcolor[HTML]{EFEFEF}-                                  & \cellcolor[HTML]{EFEFEF}-                                  \\ \hline
\multicolumn{8}{c}{\textbf{Multi-Task (Toxic + Sentiment)}}                                                                                                                                                                                                                                                                                                                                                                                                                                                        \\ \hline
\multicolumn{1}{c|}{}                                    & \multicolumn{1}{l|}{GPT2 + VM}                                     & 91.64                                             & \multicolumn{1}{l|}{91.25}                                  & -                                                          & \multicolumn{1}{c|}{-}                                                    & \multicolumn{1}{l}{76.09}                                  & \multicolumn{1}{l}{76.44}                                  \\
\multicolumn{1}{c|}{\multirow{-2}{*}{\textbf{T+V}}}      & \multicolumn{1}{l|}{RT + VM}                              & 92.33                                             & \multicolumn{1}{l|}{92.66}                                  & -                                                          & \multicolumn{1}{c|}{-}                                                    & \multicolumn{1}{l}{81.57}                                  & \multicolumn{1}{l}{81.72}                                  \\ \hline
\multicolumn{1}{c|}{}                                    & \multicolumn{1}{l|}{GPT2 + WP}                                     & 92.51                                             & \multicolumn{1}{l|}{92.34}                                  & -                                                          & \multicolumn{1}{c|}{-}                                                    & \multicolumn{1}{l}{77.73}                                  & \multicolumn{1}{l}{77.27}                                  \\
\multicolumn{1}{c|}{\multirow{-2}{*}{\textbf{T+A}}}      & \multicolumn{1}{l|}{RT + WP}                              & 93.78                                             & \multicolumn{1}{l|}{93.71}                                  & -                                                          & \multicolumn{1}{c|}{-}                                                    & \multicolumn{1}{l}{82.98}                                  & \multicolumn{1}{l}{82.64}                                  \\ \hline
\multicolumn{1}{c|}{}                                    & \multicolumn{1}{l|}{GPT2 + VM+ WP}                                 & 93.45                                             & \multicolumn{1}{l|}{93.23}                                  & -                                                          & \multicolumn{1}{c|}{-}                                                    & \multicolumn{1}{l}{79.79}                                  & \multicolumn{1}{l}{79.92}                                  \\
\multicolumn{1}{c|}{\multirow{-2}{*}{\textbf{T+V+A}}}    & \multicolumn{1}{l|}{\cellcolor[HTML]{EFEFEF}RT + VM + WP}          & \cellcolor[HTML]{EFEFEF}94.06                     & \multicolumn{1}{l|}{\cellcolor[HTML]{EFEFEF}93.94}          & \cellcolor[HTML]{EFEFEF}-                                  & \multicolumn{1}{c|}{\cellcolor[HTML]{EFEFEF}-}                            & \multicolumn{1}{l}{\cellcolor[HTML]{EFEFEF}83.05}          & \multicolumn{1}{l}{\cellcolor[HTML]{EFEFEF}82.18}          \\ \hline
\multicolumn{8}{c}{\textbf{Multi-Task (Toxic + Severity + Sentiment)}}                                                                                                                                                                                                                                                                                                                                                                                                                                             \\ \hline
\multicolumn{1}{c|}{}                                    & \multicolumn{1}{l|}{GPT2 + VM}                            & 92.72                                             & \multicolumn{1}{l|}{92.65}                                  & \multicolumn{1}{l}{85.89}                                  & \multicolumn{1}{l|}{85.59}                                                & \multicolumn{1}{l}{78.22}                                  & \multicolumn{1}{l}{77.59}                                  \\
\multicolumn{1}{c|}{\multirow{-2}{*}{\textbf{T+V}}}      & \multicolumn{1}{l|}{RT + VM}                                       & 93.51                                             & \multicolumn{1}{l|}{93.01}                                  & \multicolumn{1}{l}{85.48}                                  & \multicolumn{1}{l|}{85.76}                                                & \multicolumn{1}{l}{82.38}                                  & \multicolumn{1}{l}{82.49}                                  \\ \hline
\multicolumn{1}{c|}{}                                    & \multicolumn{1}{l|}{GPT2 + WP}                            & 92.81                                             & \multicolumn{1}{l|}{92.59}                                  & \multicolumn{1}{l}{85.97}                                  & \multicolumn{1}{l|}{85.52}                                                & \multicolumn{1}{l}{78.19}                                  & \multicolumn{1}{l}{77.38}                                  \\
\multicolumn{1}{c|}{\multirow{-2}{*}{\textbf{T+A}}}      & \multicolumn{1}{l|}{RT + WP}                                       & 93.88                                             & \multicolumn{1}{l|}{93.73}                                  & \multicolumn{1}{l}{85.91}                                  & \multicolumn{1}{l|}{86.01}                                                & \multicolumn{1}{l}{82.45}                                  & \multicolumn{1}{l}{82.66}                                  \\ \hline
\multicolumn{1}{c|}{}                                    & \multicolumn{1}{l|}{GPT2 + VM+ WP}                       & 93.72                                             & \multicolumn{1}{l|}{93.56}                                  & \multicolumn{1}{l}{86.71}                                  & \multicolumn{1}{l|}{86.39}                                                & \multicolumn{1}{l}{80.03}                                  & \multicolumn{1}{l}{80.21}                                  \\
\multicolumn{1}{c|}{\multirow{-2}{*}{\textbf{T+V+A}}}    & \multicolumn{1}{l|}{\cellcolor[HTML]{EFEFEF}\textbf{RT + VM + WP}} & \cellcolor[HTML]{EFEFEF}\textbf{94.35}            & \multicolumn{1}{l|}{\cellcolor[HTML]{EFEFEF}\textbf{94.29}} & \multicolumn{1}{l}{\cellcolor[HTML]{EFEFEF}\textbf{86.84}} & \multicolumn{1}{l|}{\cellcolor[HTML]{EFEFEF}\textbf{87.12}}               & \multicolumn{1}{l}{\cellcolor[HTML]{EFEFEF}\textbf{83.42}} & \multicolumn{1}{l}{\cellcolor[HTML]{EFEFEF}\textbf{82.43}} \\ \hline
\end{tabular}}
\end{table}

\subsection{Findings from Experiments}
Table~\ref{tab:ml_results} shows the results of toxicity, severity and sentiment classification tasks with different baseline models. Results of the proposed {\em ToxVidLM} model are shown in Table~\ref{tab:dl_results}. From all these reported results, we can conclude the following: \newline

\textbf{(1)} Our experimentation involved four text encoders (HingRoberta, HingGPT, IndicBERT, HingMBERT), two video models (VideoMAE, Timesformer), and two audio encoders (Whisper, MMS) to discern optimal performers for toxic video detection in Hindi-English code-mixed language. Analysis of the outcomes, presented in Table \ref{tab:ml_results}, reveals the superior performance of HingRoberta/HingGPT, VideoMAE, and Whisper as the best encoders for text, video, and audio modalities, respectively. These top-performing models are subsequently incorporated into our proposed framework.

\textbf{(2)} Across all three tasks, unimodal baselines demonstrate that the text modality consistently outperforms video and audio modalities. This underscores the paramount importance of text modality in toxicity detection within videos. Notably, employing the HingRoberta model for the text modality yields the highest F1 score of 86.98\% in toxicity detection, while VideoMAE and Whisper models achieve accuracy values of 68.67\% and 76.18\%, respectively, in video and audio modalities. Similar trends are observed in the other two tasks, namely severity and sentiment analysis. Hence we prioritize text modality as a base in our proposed model's performance analysis.

\textbf{(3)} In the context of bimodal baselines, the text+audio configuration consistently demonstrates superior performance compared to the other two combinations across all tasks. Notably, the most effective baseline considering three modalities (RT+VM+WP) achieves the highest accuracy of 88.08\%, 78.66\%, and 75.82\% for toxicity, severity, and sentiment tasks, respectively.

\textbf{(4)} In both single-task and multitask scenarios, our proposed model ({\em ToxVidLM}) consistently outperforms all baselines by a substantial margin. In the three-task settings, the (RT + VM + WP) variants exhibit enhancements in accuracy, surpassing the best baseline model by 6.21\%, 8.46\%, and 6.61\% for toxicity, severity, and sentiment tasks, respectively. These notable improvements underscore the effectiveness of our proposed model, attributed to the incorporation of an innovative modality synchronization module compared to baseline approaches.

\textbf{(5)} The multitask (MT) variants of our proposed model consistently surpass its single-task (ST) variant across all tasks within the same encoding setting. Notably, in a two-task scenario utilizing GPT2+VM encoders, the MT variant outperforms the ST counterpart, demonstrating improvements in $F_1$-score of 2.57\%, 2.54\%, and 2.58\% for toxicity, severity, and sentiment tasks, respectively. These results suggest that incorporating sentiment and severity knowledge enhances the performance of the toxicity detection task and contributes to overall model improvement. Conversely, no substantial improvements are observed when comparing multitask settings with two tasks versus three tasks.

{\em Statistical Analysis}: 
A statistical t-test was conducted on the values from ten runs of both the proposed models and baseline models, yielding p-values below 0.05, indicating the statistical significance of the results. The t-test was implemented using functions from the scipy library\footnote{\url{https://docs.scipy.org/doc/scipy-1.6.3/reference/generated/scipy.stats.ttest_ind.html}}. We have highlighted (gray color) the results in Table~\ref{tab:ml_results} and ~\ref{tab:dl_results} which are statistically significant.

\subsubsection{Ablation Study}
\begin{table}[hbt]
\centering
\caption{Ablation study to show the effect of gated fusion (GF), multi-head cross attention (MHCA) in proposed {\em ToxVidLM} model}
\label{Abl}
\scalebox{0.70}{
\begin{tabular}{l|ll|ll|ll}
\hline
\multirow{2}{*}{\textbf{Model}} & \multicolumn{2}{c|}{\textbf{Toxic}}                                 & \multicolumn{2}{c|}{\textbf{Severity}}                              & \multicolumn{2}{c}{\textbf{Sentiment}}                             \\ \cline{2-7} 
                                & \multicolumn{1}{c}{\textbf{F1}} & \multicolumn{1}{c|}{\textbf{Acc}} & \multicolumn{1}{c}{\textbf{F1}} & \multicolumn{1}{c|}{\textbf{Acc}} & \multicolumn{1}{c}{\textbf{F1}} & \multicolumn{1}{c}{\textbf{Acc}} \\ \hline
\textbf{ToxVidLM}                  & \textbf{94.35}                  & \textbf{94.29}                    & \textbf{86.84}                  & \textbf{87.12}                    & \textbf{83.42}                  & \textbf{82.43}                   \\
- GF                            & 92.63                           & 92.84                             & 84.11                           & 84.59                             & 80.29                           & 80.54                            \\
- MHCA                          & 89.87                           & 89.92                             & 81.45                           & 81.56                             & 77.38                           & 77.62                            \\
- MHCA - GF                     & 87.72                           & 87.86                             & 78.22                           & 78.44                             & 75.25                           & 75.48                            \\ \hline
\end{tabular}}
\end{table}
We conducted an ablation study (see Table~\ref{Abl}) on our proposed model, {\em ToxVidLM}, to elucidate the impact of gated fusion (GF) and multi-head cross attention (MHCA) in toxic video detection. The removal of GF from {\em ToxVidLM} results in a discernible decrease of 1.72\%, 2.73\%, and 3.13\% in F1-score for toxicity, severity, and sentiment tasks, respectively. This decline underscores the crucial role of the gated fusion module in effectively fusing video and audio modalities, thereby enhancing overall performance. Upon removing the MHCA component from {\em ToxVidLM}, a substantial performance drop is observed across all tasks, affirming the significant impact of MHCA in our proposed model. This proves MHCA is instrumental in generating text-guided audio and video features. Notably, the simultaneous exclusion of both GF and MHCA components results in a substantial drop of 6.63\% in the F1 score for the toxicity task, with similar performance reductions observed in other tasks. This considerable decline underscores the pivotal role of the cross-modal synchronization module, emphasizing its capacity to align representations from three distinct modalities within a unified space. Please see the qualitative analysis of our proposed framework in Appendix~\ref{sec:error}. 

\section{Conclusion and Future Works}
In an ever-evolving internet landscape, where videos have become the predominant form of content, the challenge of detecting toxic content, especially in low-resource code-mixed languages, is more critical than ever. We introduce {\em ToxCMM}, a pioneering benchmark dataset featuring code-mixed videos for toxic content detection. Our proposed LM-based advanced multimodal framework ({\em ToxVidLM}) achieved remarkable results, emphasizing the significance of combining text, audio, and video modalities. It is worth noting that, among the individual modalities, transformer encodings of text prove to be particularly effective in detecting toxic videos. Beyond toxicity, the {\em ToxCMM} dataset includes two additional labels, sentiment, and severity, offering a comprehensive resource for further exploration in sentiment analysis within low-resource code-mixed videos. Our research emphasizes AI's role in fostering a respectful online environment and promoting civility against toxic speech in video data. 

\section{Limitations}
\label{sec:limitation}
Our endeavour aimed to construct a multimodal framework and introduce a benchmark dataset, ToxCMM, tailored for detecting toxic video content within code-mixed language. However, it is crucial to acknowledge certain inherent limitations in our proposed approach and dataset, including:

1. In this study, we did not consider the context of the video clip; we treated a single utterance as the input post. Future investigations will incorporate the entire video clip as input, recognizing the pivotal role of context in deciphering the actual meaning of the utterance.

2. Implicit or indirect toxic expressions were excluded from this study, primarily focusing on explicit markers. Future work will address the development of datasets and models capable of detecting implicit/indirect toxic posts.

3. The proposed {\em ToxVidLM} model fine-tunes two encoder modules, an LM, and additional modules, requiring a considerable amount of GPU memory for training. Due to computational limitations, we were unable to experiment even with parameter-efficient fine-tuning methods (PEFT) like LoRA \cite{hu2021lora} or Quantized-LoRA \cite{dettmers2023qlora} for models containing billions of parameters like OpenHathi-7B\footnote{\href{https://huggingface.co/sarvamai/OpenHathi-7B-Hi-v0.1-Base}{https://huggingface.co/sarvamai/OpenHathi-7B-Hi-v0.1-Base}}, Airavata-7B \cite{gala2024airavata}, Llama 2-7B \cite{touvron2023llama}, or Mistral-7B \cite{jiang2023mistral}. However, since our model is versatile, those with ample GPU resources can easily substitute larger models into the textual side, potentially achieving enhanced performance specifically for video classification tasks.

\bibliography{anthology,custom}
\bibliographystyle{acl_natbib}

\appendix

\section{Qualitative Analysis}
\label{sec:error}
\begin{figure*}[h]
	\centering
        \includegraphics[width=\textwidth]{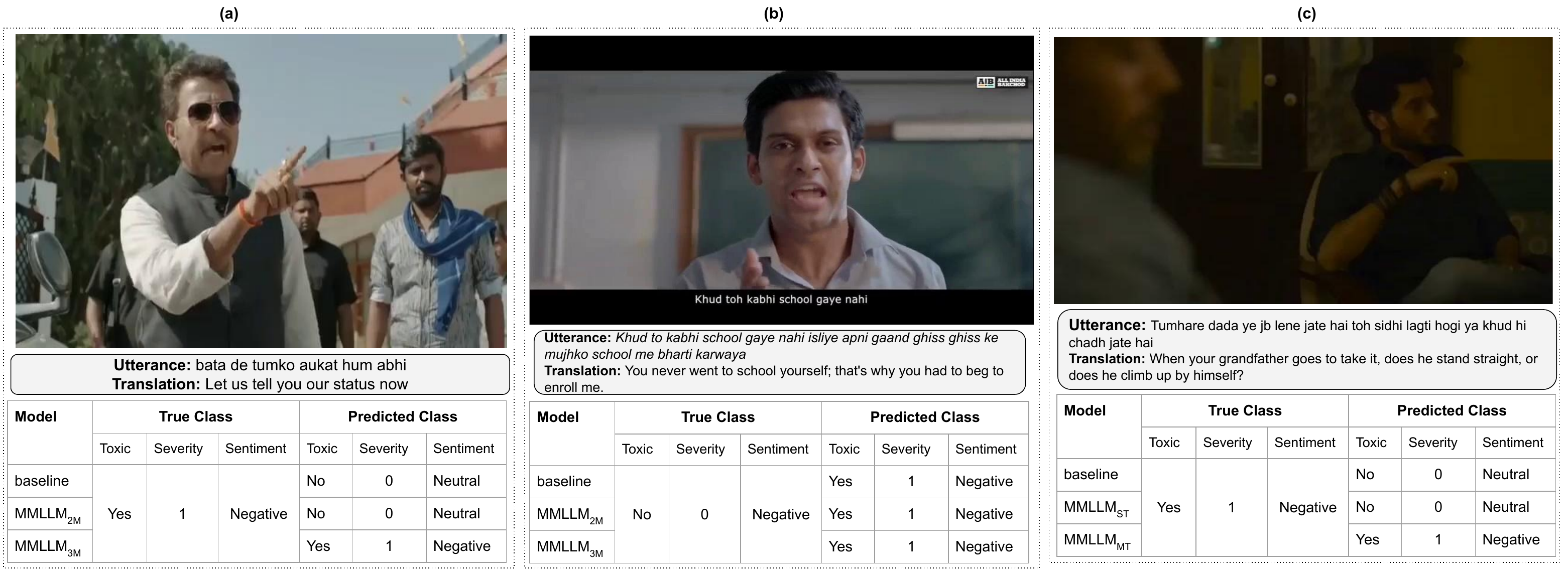}
	\caption {Human annotation versus model’s prediction for qualitative analysis with different settings}
	\label{fig:error}
	
\end{figure*}

Figure~\ref{fig:error} illustrates qualitative results comparing the insightfulness of various tasks between ground truth and model predictions.

(i) In the first sample (a), the textual modality ("Let us tell you our status now") lacks offensive words but constitutes a toxic utterance with negative sentiment and a severity score of 1 based on acoustic and visual expression. The user is threatening someone. Both the best baseline model and our proposed model with two modalities settings (ToxVidLM$_{2M}$) make incorrect predictions across all tasks. However, considering three modalities variants (ToxVidLM$_{3M}$) allows accurate prediction of all classes. This observation underscores our proposed model's enhanced comprehension of diverse modalities, demonstrating that incorporating audio and visual cues with text provides a superior understanding of video data.

(ii) The second example's true labels are non-toxic with negative sentiment and a severity score of 0. Regrettably, both baseline and proposed models mispredict toxicity and severity labels. Although the surface sentiment appears negative, and some negative words and angry facial expressions are present, understanding the actual implicit meaning requires knowledge of the video clip's previous context. Since this study focuses on stand-alone utterance labels without considering previous context, all models make incorrect predictions. Integrating context represents a potential future direction for this work.

(iii) In the third example, both baseline and single-task variants of the proposed model (ToxVidLM$_{ST}$) inaccurately predict all labels, while the multitask model (ToxVidLM$_{MT}$) correctly identifies all classes. This example is toxic due to offensive language and derogatory assumptions about someone's grandfather. The terms "sidhi lagti hogi" and "khud hi chadh jate hai" are disrespectful, implying negative sentiment. Task-specific layers in the multitask framework aid in correctly identifying true sentiment, leading to the accurate identification of toxicity and severity labels in the given post.

\section{Encoder Description}
\label{sec:encoder}

\textbf{Audio Encoders:}
We conducted experiments using two state-of-the-art (SOTA) models, namely Whisper~\citet{radford2023robust} and MMS\cite{pratap2023scaling}, to encode audio signals and extract meaningful representations from the audio data. In our investigation, Whisper consistently demonstrated superior performance across all experimental settings compared to MMS. The reasons behind Whisper's superiority could be manifold. It might possess a more refined architecture tailored for audio processing, incorporating domain-specific optimizations or leveraging advanced techniques such as self-attention mechanisms or convolutional layers optimized for audio data. Additionally, the training procedure, hyperparameter settings, or data preprocessing techniques employed for Whisper could contribute to its superior performance over MMS.

\textit{(i) Massively Multilingual Speech (MMS)~\cite{pratap2023scaling}: } Developed by Facebook AI, MMS is a comprehensive multilingual pre-trained model for speech. It undergoes pretraining with Wav2Vec2's self-supervised training objective on an extensive dataset comprising approximately 500,000 hours of speech data across more than 1,400 languages. 

\textit{(ii) Whisper: }~\citet{radford2023robust} introduced Whisper, a novel multilingual speech recognition model. Whisper is trained on an extensive audio dataset, incorporating weak supervision for improved performance.

\textbf{Video Encoders:}
We explored the efficacy of two transformer-based video models equipped with spatiotemporal context by uniformly sampling 16 frames from each video clip and feeding them into these encoders. Our experiments focused on comparing the performance of VideoMAE ~\citet{tong2022videomae} and TimeSformer ~\cite{bertasius2021space}, with VideoMAE consistently outperforming TimeSformer across all settings.

Despite the advanced design and capabilities of TimeSformer, our experimental results indicate that VideoMAE consistently outperforms it across all evaluated settings. The reasons behind this superiority could stem from various factors such as the efficiency of VideoMAE's learning approach, its ability to capture subtle temporal dependencies, or the effectiveness of its feature representation in downstream tasks. Additionally, factors like model architecture, training strategies, and hyperparameter settings may also influence the comparative performance of these models.

\textit{(i) VideoMAE: }~\citet{tong2022videomae} introduced a data-efficient learning approach for self-supervised video pre-training. It utilizes Masked Autoencoders to efficiently learn representations from video data, demonstrating effectiveness in enhancing model performance.

\textit{(ii) TimeSformer: }~\cite{bertasius2021space} This model is a novel architecture designed for video understanding tasks. It extends the Transformer architecture to capture temporal relationships in videos by incorporating a spatiotemporal attention mechanism, demonstrating state-of-the-art performance in various video analysis applications.

\textbf{Text Encoders:}
 \cite{nayak2022l3cube} introduces several transformer-based models pre-trained on L3Cube-HingCorpus, the first large-scale real Hindi-English code-mixed dataset in Roman script: HingBERT, HingMBERT, HingRoBERTa, and HingGPT. These models, evaluated on tasks like sentiment analysis, POS tagging, NER, and language identification, demonstrate the effectiveness of real code-mixed data. They also present HingBERT-LID for language identification and HingFT for code-mixed word embeddings, making all resources publicly available for further research in Hinglish NLP.\\
\textit{(i) HingBERT}:
    HingBERT is a BERT-based model pre-trained on the Hinglish corpus using masked language modeling objectives. It is evaluated on downstream tasks such as code-mixed sentiment analysis, POS tagging, NER, and language identification (LID) from the GLUECoS benchmark.
    
\textit{(ii) HingMBERT}:
    HingMBERT is a variant of the multi-lingual BERT model pre-trained on the Hinglish corpus. It is trained using both Roman and Devanagari scripts and is assessed on various downstream tasks including code-mixed sentiment analysis, POS tagging, NER, and LID.
    
\textit{(iii) HingRoBERTa}: 
    HingRoBERTa is a RoBERTa-based model trained on the Hinglish corpus. It has versions trained on Roman script and a combination of Roman + Devanagari scripts. The model is evaluated on downstream tasks such as code-mixed sentiment analysis, POS tagging, NER, and LID.

\textit{(iv) HingGPT}: 
    HingGPT is a generative transformer model based on the GPT-2 architecture. It is trained on the Hinglish corpus to generate full tweets in code-mixed Hinglish.

We initially utilized all four pre-trained models for the unimodal baselines and filtered the top 2 models namely HingGPT and HingRoBERTa for the rest of the experiments. According to the results obtained from our main framework, HingRoBERTa outperforms HingGPT in all the corresponding settings, proving it to be the best candidate for our framework among all the considered models.

\end{document}